\documentclass{ceurart}

\usepackage{algorithm}
\usepackage{algorithmic}
\usepackage{listings}
\usepackage[acronym]{glossaries}
\newacronym{cea}{CEA}{cell entity annotation}
\newacronym{cot}{CoT}{chain-of-thought}
\newacronym{dpo}{DPO}{direct preference optimization}
\newacronym{gpu}{GPU}{graphics processing unit}
\newacronym{irpo}{IRPO}{iterative reasoning preference optimization}
\newacronym{llm}{LLM}{large language model}
\newacronym{ln}{LN}{length normalization}
\newacronym{lora}{LoRA}{low-rank adaptation}
\newacronym{lrpo}{L-RPO}{length-normalized regularized preference optimization}
\newacronym{rpo}{RPO}{regularized preference optimization}
\newacronym{sft}{SFT}{supervised fine-tuning}

\glsdisablehyper

\begin{document}

\copyrightyear{2026}
\copyrightclause{Copyright for this paper by its authors.
  Use permitted under Creative Commons License Attribution 4.0
  International (CC BY 4.0).}
\conference{OM 2026: The 21st International Workshop on Ontology Matching collocated with the 25th International Semantic Web Conference (ISWC 2026), October 25th, 2026, Bari, Italy}

\title{TELLER: Dual-Path Iterative Preference Optimization for Table Entity
Linking}

\author[1]{Kehao Li}[
  orcid=0009-0002-7704-1049]
\fnmark[1]
\address[1]{RWTH Aachen University, Aachen, Germany}
\author[1]{Yixin Peng}[
  orcid=0009-0003-9777-0123,
  email=peng@dbis.rwth-aachen.de]
\cormark[1]
\fnmark[1]
\author[1]{Stefan Decker}[
  orcid=0000-0001-6324-7164]
\cortext[1]{Corresponding author.}
\fntext[1]{These authors contributed equally.}

\begin{abstract}
  Entity linking in tables matches short and ambiguous cell mentions to their
  corresponding knowledge base entities. Existing approaches typically rely on data preprocessing pipelines that retain either compact or extensive table content as contextual evidence, 
  and then formulate entity linking as a language generation task for instruction-tuned models; recent approaches further incorporate explicit reasoning to disambiguate challenging mentions. 
  However, their training supervision is usually static: fixed preference data cannot adapt to the residual errors of an
  evolving model, while variations in reasoning length can bias sequence-level preference learning. To address these limitations, we present
  TELLER: Table Entity Linking through Learning from Errors and Reasoning. We first retrieve and rank
  Wikidata candidates and retain reduced table evidence in the prompt. 
  The direct-answer path applies iterative direct preference optimization and
  refreshes its preference data with residual errors from the updated model.
  The reasoning path uses filtered and compressed chain-of-thought rationales
  for supervised fine-tuning, followed by our iterative length-normalized
  regularized preference optimization.
  On the TableInstruct entity-linking subset, the direct-answer path improves
  accuracy from 94.35\% to 94.50\%; on the MammoTab V2
  evaluation set, it improves accuracy from 87.59\% to 88.20\%. The reasoning
  path improves accuracy from 92.90\% to 92.95\% on TableInstruct and from
  79.09\% to 81.85\% on MammoTab V2, while maintaining high rates of complete reasoning generation. These results show that
  iterative preference learning benefits both concise entity prediction and
  explicit reasoning.
\end{abstract}


\maketitle

\section{Introduction}
Tables are a major source of structured knowledge, yet their cell values are
often too short and ambiguous to be interpreted in isolation.  In
\gls{cea}, each target mention must be linked to its corresponding
knowledge base entity.  The decisive evidence may occur in a column header or
table caption rather than in the mention itself, while unrelated cells can
introduce competing entities and distract the linker~\cite{katsakioris2022entity}.
Effective table entity linking therefore depends on two closely related design
choices: which candidate entities are presented to the model and how much
table context is retained in the input.

Recent work treats entity linking as a language generation task.  GENRE directly
generates canonical entity names~\cite{decao2021autoregressive}, while
INSGENEL~\cite{xiao2023instructed} and EntGPT~\cite{ding2024entgpt} use
instruction-tuned language models for this task.  For tables, TableLlama~\cite{zhang2023tablellama} shows
that one instruction format can support many table-understanding
tasks.  These studies motivate us to formulate
\gls{cea} as candidate-conditioned generation.  The input contains a target
cell, selected table context, and candidate entities described by their names,
types, and descriptions.  The model selects the candidate that best matches
the evidence in the input.  Building large datasets in this format, however,
requires a consistent way to retrieve candidates and serialize tables.  The serialized input should preserve the evidence needed to distinguish among
the candidates, exclude unrelated cells that may distract the model, and keep
the input length manageable.

Explicit reasoning can improve language-generation models by guiding them how
to work through a problem step-by-step.  For example, \gls{cot}
supervision~\cite{wei2022chain} can make candidate comparisons and reasoning steps visible in the
output.  This makes a prediction easier to interpret and may help the model
select the correct entity.  Since collecting human-written rationales is difficult and expensive, large
language models can instead be used to generate rationale supervision for
smaller models~\cite{wiegreffe-etal-2022-reframing,li-etal-2023-symbolic}.  However,
teacher-generated rationales may contain unsupported claims or repeated
content.  They may also be too long for a smaller model to complete within its
generation budget.

Preference optimization provides another way to improve a model by learning
from the entity-linking errors that remain after \gls{sft}.  Each training
example has a gold entity, so an incorrect model prediction can serve as the
rejected response and the gold answer as the chosen response.  This procedure
requires neither human preference labels nor a \gls{llm} judge.
\Gls{dpo}~\cite{rafailov2023direct} can learn from these pairs, but a fixed set
of pairs captures only the errors made by the model that generated them.  After
the model is updated, its remaining errors are more relevant for the next
update~\cite{xiong2024iterative}.  Applying preference optimization to reasoning
outputs introduces two further problems.  First, harder entity-linking
examples often require longer rationales, and differences in response length
can bias sequence-level preference
scores~\cite{park2024disentangling,meng2024simpo}.  Second, increasing only the
relative margin between the chosen and rejected responses does not guarantee
that the gold-consistent chosen rationale remains likely under the model.

To address these limitations, we propose TELLER.  Its offline preprocessing pipeline retrieves and ranks plausible Wikidata~\cite{vrandecic2014wikidata}
candidates. The prompt retains the target cell, the other cells in its row and
column, the table headers, and the caption, while removing all remaining table
cells. This compact representation preserves useful context
while reducing interference from non-target entities.  TELLER then improves
entity linking through two iterative preference-learning paths.  The
direct-answer path applies \emph{iterative DPO} after SFT.  At each iteration,
the current model generates predictions on the training data.  Its incorrect
predictions are paired with the corresponding gold answers, and these new pairs
are used for the next DPO update.

The reasoning path first uses filtered and compressed teacher rationales for
CoT-SFT and then applies \emph{iterative L-RPO}. We introduce L-RPO for preference pairs containing unequal-length reasoning
responses. L-RPO retains the chosen-response likelihood
regularizer introduced by RPO~\cite{liu2024regularized}, but extends and modifies its
pairwise preference term by replacing summed sequence log-probabilities with
per-token average log-probabilities for both the trainable policy and the
frozen reference model. This length-normalized, reference-relative preference
margin is the key contribution of L-RPO. We further apply L-RPO
iteratively: after each round, the updated model generates new responses, and
its residual reasoning errors replace the previous rejected responses in the
next-round preference dataset.
Both iterative methods can therefore continue for multiple rounds.  In this
work, we run two rounds of each method to provide a controlled comparison and
demonstrate the effect of refreshing the preference data.  The two paths also
allow us to study concise entity predictions separately from longer
reasoning-and-answer sequences.

The two paths are evaluated on both the entity-linking subset of TableInstruct~\cite{zhang2023tablellama}
and the MammoTab V2 evaluation set~\cite{cremaschi2025mammotab}.  Iterative DPO produces progressive accuracy gains over direct-answer
SFT: 94.35\% $\rightarrow$ 94.40\% $\rightarrow$ 94.50\% on TableInstruct and
87.59\% $\rightarrow$ 88.16\% $\rightarrow$ 88.20\% on MammoTab V2.  The
final MammoTab score of 0.882 exceeds the best baseline result of 0.86 listed
on the public MammoTab V2 evaluation page~\cite{mammotabV2evaluation}.  In the reasoning
path, iterative L-RPO improves CoT-SFT accuracy from 92.90\% to 92.95\% on
TableInstruct and from 79.09\% to 81.85\% on MammoTab V2.  Complete-rationale
rates reach 99.55\% and 91.86\%, respectively.  Direct-answer models remain
more accurate, showing that explicit reasoning introduces a measurable cost
even when preference optimization recovers part of the gap.

The main contributions of this work are:
\begin{itemize}
  \item We develop an offline pipeline for retrieving and ranking Wikidata
  candidates and introduce a compact prompt serialization that retains the table
  headers, caption, and cells in the target cell's row and column while removing
  all remaining table cells.

  \item We introduce iterative DPO for direct-answer entity linking.  Starting
  from the best SFT checkpoint, each iteration constructs preference pairs
  from the current model's errors and uses them for the next DPO update.

  \item We propose L-RPO for reasoning-oriented preference learning by
  combining completion-length normalization with chosen-response
  regularization.  Its iterative extension can repeatedly refresh reasoning
  preference pairs with the residual errors of the evolving L-RPO model after
  CoT-SFT.

  \item We evaluate the complete training pipeline on the TableInstruct
  entity-linking subset and MammoTab V2.  Iterative DPO reaches 94.50\% and
  88.20\% accuracy, respectively; the latter corresponds to 0.882 CEA and
  exceeds the best result currently listed on the public benchmark.  Our
  stage-wise analysis also quantifies the benefit of iterative L-RPO for
  CoT-SFT.
\end{itemize}

\section{Related Work}

\subsection{Generative Entity Linking and Table Understanding}
Entity linking has typically been decomposed into candidate retrieval and
disambiguation.  BLINK, for example, uses a bi-encoder to retrieve entities and
a cross-encoder to rerank them~\cite{wu2020scalable}.  Generative methods replace
or augment this ranking stage with sequence generation: GENRE performs
constrained autoregressive generation over canonical entity names
~\cite{decao2021autoregressive}, while INSGENEL combines a lightweight retriever
with an instruction-tuned generator~\cite{xiao2023instructed}.  EntGPT further
studies prompted and instruction-tuned  \gls{llm} for entity
linking~\cite{ding2024entgpt}.  Recent work also considers fine-tuning-free
few-shot linking and adaptive routing that reserves explicit reasoning for
difficult mentions~\cite{liu2024onenet,li2025leveraging}.  These approaches show
that generation offers a flexible task interface, but they do not remove the
need for high-recall candidate construction or informative mention context.

Both requirements become more acute for tables, where an isolated cell often
contains little lexical evidence.  TURL introduces structure-aware
pre-training over table metadata, headers, and entities~\cite{deng2020turl}, and
RoCEL explicitly models row context as relational evidence and column context
as categorical evidence~\cite{wang2024rocel}.  Complementary analyses show that
headers, captions, and structurally related cells are useful, whereas allowing
attention to unrelated cells can dilute the relevant signal
~\cite{katsakioris2022entity}.  At a broader level, TableLlama demonstrates that a single instruction format can support heterogeneous table
tasks~\cite{zhang2023tablellama}.  Together, these results motivate selective
serialization rather than treating a table as either an isolated cell or an
undifferentiated token sequence.

TELLER builds on these two strands by using a generator as a
candidate-conditioned discriminator.  Each retrieved Wikidata candidate is
represented by its name, type, and description, while the prompt retains
compact metadata and structurally relevant table evidence.  The candidate set
anchors the output to the target knowledge base, and the instruction format
provides a common interface for our training dataset.  Our focus is a controlled input and training formulation that makes
the same entity-linking instances usable for direct prediction, rationale
supervision, and iterative preference learning.

\subsection{Rationale Supervision and Distillation}
Chain-of-thought (CoT) prompting elicits intermediate reasoning from
in-context demonstrations~\cite{wei2022chain}; rationale distillation instead
places teacher-generated reasoning in the student's training target.
Fine-tune-CoT established this teacher--student recipe~\cite{ho2023large}, and
subsequent work showed that rationales can transfer useful supervision to much
smaller models or reduce the amount of labeled data required
~\cite{magister2023teaching,hsieh2023distilling}.  Multi-path and
self-consistent variants seek better targets by aggregating or contrasting
alternative rationales~\cite{chen2023mcckd,wang2023scott}.  Related
self-training methods such as STaR retain reasoning traces that lead to correct
answers and regenerate unsuccessful traces before another training round
~\cite{zelikman2022star}.

Teacher generation, however, does not guarantee reliable supervision.  CoT
explanations can be inconsistent with the model's actual decision process
~\cite{turpin2023language}, and even answer-consistent traces may contain
unsupported or unhelpful steps.  DOCTOR therefore treats the teacher as
unreliable and selectively distills rationales that pass consistency and
helpfulness filters~\cite{chae2023dialogue}.  Efficiency is a second concern:
long or repetitive traces increase training and decoding costs without
necessarily adding useful signal.  Keypoint-based distillation emphasizes
salient rationale tokens~\cite{feng2024keypoint}; controlled studies likewise
find that a small subset of key tokens can retain much of the benefit of
CoT-augmented distillation~\cite{wadhwa2024mysteries}.  TokenSkip instead learns
to omit less informative reasoning tokens at generation time
~\cite{xia2025tokenskip}.

Our rationale pipeline follows the same quality-over-quantity principle but
specializes it to candidate comparison in table entity linking.  We require
the final prediction to agree with the gold entity, reject malformed,
unsupported, leaked, or repetitive traces, and extractively shorten overlong
rationales without changing their conclusions.  The retained rationale is an
explicit CoT-SFT target grounded in the supplied table and candidate evidence;
we do not treat fluency or answer correctness alone as evidence of
faithfulness.  These filtered targets subsequently serve as chosen responses
for reasoning-oriented preference optimization.

\subsection{Preference Optimization and Iterative Alignment}
DPO optimizes a language model directly from chosen--rejected response pairs
relative to a frozen reference model, avoiding both a learned reward model and
online reinforcement learning~\cite{rafailov2023direct}.  Its simplicity also
exposes two issues that matter for reasoning sequences.  First, improving the
chosen--rejected margin need not preserve the absolute likelihood of the
chosen response.  RPO addresses this failure mode by adding a supervised
negative-log-likelihood term on the chosen completion
~\cite{liu2024regularized}.  Second, summed sequence log-probabilities entangle
response quality with length and can induce unintended length preferences
~\cite{park2024disentangling}.  SimPO uses average per-token log-probability as
its implicit reward, but removes the reference model altogether
~\cite{meng2024simpo}.

A fixed offline preference set creates a separate limitation: it records the
errors of the policy that produced it, even after training has changed that
policy.  Iterative preference learning instead samples new responses from the
updated model and rebuilds the comparison data for the next round
~\cite{xiong2024iterative}.  IRPO extends this idea to reasoning by repeatedly
generating reasoning paths, selecting them by answer correctness, and training
with a DPO objective~\cite{pang2024iterative}.  Thus, iteration,
chosen-response regularization, and length control have each proved useful,
but they solve distinct problems: stale negatives, likelihood degradation,
and sequence-length bias.

TELLER addresses these problems for table entity linking. In the direct-answer path, the gold entity and a current model error form each
preference pair, and every round replaces resolved errors with the model's
residual errors.  In the reasoning path, L-RPO retains a round-specific frozen
reference model, normalizes both policy and reference scores by completion
length inside the pairwise margin, and regularizes the likelihood of the
filtered chosen rationale.  Its iterative extension regenerates rejected
reasoning responses after each update.

\section{Method}

\subsection{Task Formulation}

In this work, we focus on relational tables for entity linking. For each target cell mention \(m_i\), the table context is represented as
\[
T_i = (a_i, H_i, r_i, q_i),
\]
where \(a_i\) is the table caption, \(H_i\) is the ordered list of column
headers, \(r_i\) is the complete row containing the target cell, and \(q_i\)
contains the cells in the target column from the remaining rows. The document
context is represented as
\[
P_i = (p_i, s_i),
\]
where \(p_i\) and \(s_i\) are the Wikipedia page title and section title,
respectively. Here, ``section context'' refers only to the section title, not
to the full section text. Captions, headers, page titles, and section titles
are therefore retained as distinct fields rather than merged into an
unstructured bag of words. Each instance also contains a candidate set
\(\mathcal{C}_i = \{c_{i1}, \ldots, c_{iK}\}\), where each candidate contains
an entity name, description, and semantic type. Before being passed to the model, these structured fields are deterministically
flattened into a single plain-text prompt \(x_i\). The model therefore receives
neither a JSON object nor an unordered set of words. Missing metadata fields,
such as an unavailable caption or section title, are omitted from the
corresponding textual segment. The goal is to predict the gold entity \(e_i\) for the target cell mention. In the reasoning-based stages, the model will generate the entity prediction in the required output format with reasoning between the \verb|<think>| and \verb|</think>| tags. During evaluation, we extract the final entity prediction, and a prediction is correct only when it exactly matches the gold entity. We do not use fuzzy matching or semantic correction.

\subsection{Framework Overview}
Our framework consists of two stages: data construction and model training. 
First, we convert the original MammoTab V2 data into the TableInstruct format. 
After removing duplicate training examples and ensuring that the training set does not overlap with the validation or test sets, we merge the processed MammoTab V2 data with TableInstruct to construct direct-answer, reasoning, and preference datasets. 
For model training, we compare two paths: direct-answer SFT and reasoning-enhanced training (CoT-SFT). 
The direct-answer path performs iterative DPO following SFT. And the reasoning-enhanced path continues training from the same SFT checkpoint using CoT-SFT, followed by iterative L-RPO. 

\subsection{Data Construction}

\paragraph{Entity-linking instances.}

TableInstruct already includes entity-linking examples with a list of candidate entities. For each target cell in MammoTab, we retain only the target cell mention, the other cells in the same row and
column, table headers, caption, page title, and section title, while removing all remaining table cells. 
This shorter representation reduces the input length and limits distraction from unrelated entities in nearby cells, while preserving the more useful header and caption information. An example of the processed entity-linking data is provided in Appendix~\ref{appendix:el-example}.
To retrieve candidate entities, we build a local Wikidata index from the \texttt{latest-truthy.nt.bz2} dump \cite{vrandecic2014wikidata,wikidataDatabaseDownload}. 
We first retrieve candidates through alias matching and BM25 search \cite{robertson2009probabilistic}. 
We then rank the retrieved entities using a weighted combination of five signals, as detailed in Algorithm~\ref{alg:candidate-ranking}: name similarity, mention--entity frequency, type compatibility, BM25 relevance, and context overlap. The top $20$ candidates are retained for each target mention. Each candidate is serialized as follows: \texttt{\textless entity name [DESCRIPTION] description [TYPE] semantic type\textgreater}.


\begin{algorithm}
\caption{Candidate Retrieval and Scoring for Entity Linking}
\label{alg:candidate-ranking}
\begin{algorithmic}[1]
\REQUIRE mention $m$, column name $c$, table context $\mathrm{ctx}$ (page/section/row text),
         mention-to-entity prior counts $\mathrm{Prior}$, top-$k$ (default $20$)
\ENSURE top-$k$ ranked candidate entities
\STATE $\hat m \leftarrow \textsc{Normalize}(m)$
\STATE $E \leftarrow \textsc{RetrieveCandidates}(\hat m)$ \COMMENT{exact alias match $\cup$ BM25 full-text search over local KB}
\FOR{each entity $e \in E$}
    \STATE $f_{\mathrm{lex}}(e) \leftarrow$ lexical match score \COMMENT{exact / substring / string-similarity between $\hat m$ and $e$'s label}
    \STATE $f_{\mathrm{prior}}(e) \leftarrow$ normalized prior $\mathrm{Prior}[\hat m][e]$ \COMMENT{how often this mention resolves to $e$}
    \STATE $f_{\mathrm{type}}(e) \leftarrow$ type-compatibility score between $c$ and $e$'s entity type
    \STATE $f_{\mathrm{ctx}}(e) \leftarrow$ context overlap between $e$'s description and $\mathrm{ctx}$
    \STATE $f_{\mathrm{bm25}}(e) \leftarrow$ normalized BM25 relevance score
    \STATE $\mathrm{score}(e) \leftarrow \sum_i w_i \cdot f_i(e)$ \COMMENT{fixed heuristic weights, $i \in \{\mathrm{lex, prior, type, ctx, bm25}\}$}
\ENDFOR
\STATE $E_{\mathrm{sorted}} \leftarrow \textsc{SortDescending}(E, \mathrm{score})$
\STATE \textbf{return} top-$k$ entities of $E_{\mathrm{sorted}}$
\end{algorithmic}
\end{algorithm}

If the gold entity is not retrieved, we do not manually insert it into the candidate list. But we find that every data in our final dataset included its gold entity among the retrieved candidates. 
Therefore, no training data has an empty or NIL target in the final dataset. 
Moreover, to reduce the number of training examples that share nearly identical table contexts, we randomly sample one target cell from each table with random seed $42$. 
The final SFT dataset contains 105,000 examples: 35,000 from TableInstruct and 70,000 processed examples from MammoTab.

\paragraph{Reasoning data construction.}

We add a rationale to each data in the merged dataset and place it between the \texttt{\textless think\textgreater} and \texttt{\textless/think\textgreater} tags. 
DeepSeek-V4-Pro generates rationales for the MammoTab examples \cite{deepseekai2026deepseekv4}, while GPT-5.2 Thinking generates rationales for the TableInstruct examples \cite{openai2025gpt52}. This generation step yields 36,276 outputs.
We then apply a task-specific filter. We remove examples with invalid formats, unsupported claims, repeated content, answer leakage, or a final entity prediction that differs from the gold entity. 
For outputs longer than 500 tokens, we remove copied candidate lists, repeated reasoning, and sentences that do not provide useful evidence. We do not rewrite the remaining sentences or change the final entity prediction. 
After filtering, the CoT-SFT dataset contains 35,641 examples. The training pipeline performs a final check of sequence length and target correctness, leaving 35,617 examples for training.

\paragraph{Direct-answer iterative preference supervision.}
Let \(\pi_{0}\) be the direct-answer model selected on the validation set after
SFT.  In round \(k\), the current model \(\pi_{k-1}\) is rolled out on the
training inputs.  For every incorrect prediction, the model response becomes
the rejected completion and the gold entity sequence becomes the chosen
completion.  DPO round 1 is trained on errors from \(\pi_{0}\).  The resulting
policy \(\pi_{1}\) is then rolled out again, and only its residual errors are
used to construct the second-round data.  This refresh prevents DPO round 2
from being trained on errors that the updated direct-answer model has already
resolved. An example is provided in Appendix~\ref{appendix:direct-answer-pref-example}.

\paragraph{Reasoning iterative preference supervision.}
Let \(\pi_{0}\) be the model after CoT-SFT. In this work, we conduct two rounds of L-RPO, constructing a new preference dataset in each round. In round \(k \in \{1,2\}\), the current model \(\pi_{k-1}\) generates responses for the training data using greedy decoding. 
We keep the examples for which the final entity prediction is incorrect. 
The full generated response, including the reasoning trace and the incorrect final entity, is used as the rejected response \(y_{i,k}^{-}\). 
For the chosen response \(y_{i,k}^{+}\), we first search the constructed CoT-SFT dataset for a response to the same example whose final entity matches the gold entity. 
If no valid response is found, DeepSeek-V4-Pro generates a new one. In round 1, we reuse 924 chosen responses and generate 4,052 new responses. 
In round 2, we reuse 1,040 chosen responses and generate 2,541 new responses. 
The preference dataset for round \(k\) is defined as
\begin{equation}
    \mathcal{D}_{\mathrm{pref}}^{(k)}
    =
    \left\{
    \left(x_i, y_{i,k}^{+}, y_{i,k}^{-}\right)
    \right\}_{i=1}^{N_k},
    \qquad k \in \{1,2\}.
\end{equation}

An example is provided in Appendix~\ref{appendix:cot-pref-example}.
After training on \(\mathcal{D}_{\mathrm{pref}}^{(1)}\), we use the resulting model \(\pi_{1}\) to build the second-round dataset. 
We do not reuse the incorrect outputs generated by the CoT-SFT model. Therefore, \(\mathcal{D}_{\mathrm{pref}}^{(2)}\) focuses on the errors that remain after the first L-RPO round. 
In both rounds, we keep only valid pairs whose chosen response gives the correct entity. We remove responses with invalid formats, repeated or low-quality reasoning, or leaked answers, without changing the final entity.

\subsection{Progressive Training}

\paragraph{Supervised fine-tuning.}
We use Llama 3.1 8B as the base model \cite{grattafiori2024llama3} and apply LoRA adapters \cite{hu2022lora} to the query, key, value, and output projection layers in each attention block. Given an input prompt \(x_i\) and its gold entity sequence \(e_i\), SFT minimizes the following loss:
\begin{equation}
    \mathcal{L}_{\mathrm{SFT}}
    =
    -\frac{1}{N}
    \sum_{i=1}^{N}
    \sum_{t=1}^{|e_i|}
    \log p_{\theta}
    \left(
        e_{i,t}
        \mid
        x_i, e_{i,<t}
    \right).
\end{equation}

The loss is computed only on the response tokens, while the prompt tokens are ignored. We remove examples if the full response cannot fit within the maximum sequence length. This stage teaches the model to understand the table content, candidate format and select the correct entity.

\paragraph{Chain-of-Thought Supervised Fine-Tuning.}
Starting from the SFT checkpoint with the best validation performance, we continue training the same LoRA adapter on responses that include CoT reasoning. For each example, the target response is
\begin{equation}
    z_i
    =
    \texttt{\textless think\textgreater}\;
    r_i\;
    \texttt{\textless/think\textgreater}\;
    e_i\;
    \texttt{\textless eos\textgreater},
\end{equation}
where \(r_i\) is the filtered and shortened teacher reasoning, and \(e_i\) is the gold entity output. We use teacher forcing and compute the loss only on the response tokens:
\begin{equation}
    \mathcal{L}_{\mathrm{CoT}}
    =
    -\frac{1}{N}
    \sum_{i=1}^{N}
    \sum_{t=1}^{|z_i|}
    \log p_{\theta}
    \left(
        z_{i,t}
        \mid
        x_i, z_{i,<t}
    \right).
\end{equation}

This stage teaches the model to produce reasoning steps before giving the final entity, instead of giving only a direct answer.

\begin{figure}[t]
    \centering
    \includegraphics[width=\columnwidth]{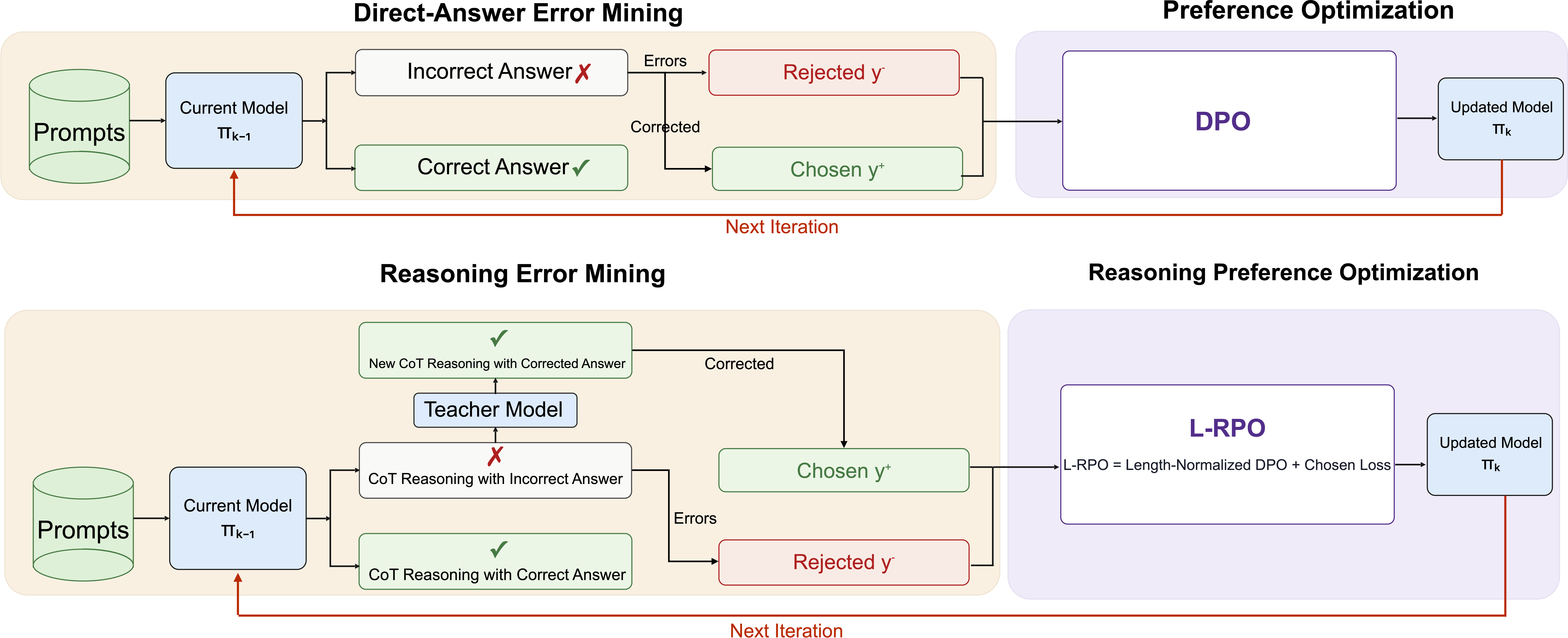}
    \caption{\textbf{Overview of the two training paths.}
    The direct-answer path starts from the SFT checkpoint with the best validation performance and applies iterative DPO. The reasoning path continues the same SFT checkpoint with
    CoT-SFT and then applies iterative L-RPO. In both paths, the model
    obtained after round $k-1$ is used to collect residual errors and construct
    fresh preference pairs for round $k$.}
    \label{fig:architecture}
\end{figure}

\paragraph{DPO and Iterative DPO.}
The direct-answer branch in Figure~\ref{fig:architecture} starts from the SFT
checkpoint with the highest validation performance. In round \(k\), the trainable
model \(\pi_{\theta,k}\) is initialized from the model obtained in the
previous stage, denoted by \(\pi_{k-1}\). For this path, \(\pi_{0}\) is the
selected SFT checkpoint. At the beginning of each round, we create a frozen
copy of \(\pi_{k-1}\) and use it as the reference model
\(\pi_{\mathrm{ref},k}\). The reference model remains fixed throughout
the round.

For a response \(y\), its sequence log-probability is the sum of the
log-probabilities of all response tokens:
\begin{equation}
    \log \pi(y \mid x)
    =
    \sum_{t=1}^{|y|}
    \log \pi
    \left(
        y_t \mid x, y_{<t}
    \right).
    \label{eq:sequence_logprob}
\end{equation}

Given a prompt \(x\), a chosen response \(y^{+}\), and a rejected response
\(y^{-}\), the standard DPO objective~\cite{rafailov2023direct} for round
\(k\) is
\begin{equation}
\begin{aligned}
    \mathcal{L}_{\mathrm{DPO}}^{(k)}
    =
    -\mathbb{E}_{(x,y^{+},y^{-}) \sim
    \mathcal{D}_{\mathrm{pref}}^{(k)}}
    \log \sigma
    \Bigg(
    \beta \Big[
        &\log \pi_{\theta,k}(y^{+} \mid x)
        - \log \pi_{\mathrm{ref},k}(y^{+} \mid x) \\
        &-
        \left(
        \log \pi_{\theta,k}(y^{-} \mid x)
        - \log \pi_{\mathrm{ref},k}(y^{-} \mid x)
        \right)
    \Big]
    \Bigg).
    \label{eq:standard_dpo}
\end{aligned}
\end{equation}

The chosen response is the gold answer, while the rejected response is an
incorrect answer generated by the current model. The DPO objective increases
the relative preference for the gold response over the incorrect response.
After DPO-k-1, we use the updated model to generate a new set of responses. Its
remaining errors are paired with the gold responses to construct fresh
preference data for DPO-k. 

\paragraph{L-RPO and Iterative L-RPO.}
The reasoning branch in Figure~\ref{fig:architecture} starts from the CoT-SFT
checkpoint and applies iterative L-RPO. In round \(k\), the trainable
model \(\pi_{\theta,k}\) is initialized from the model obtained in the
previous stage, \(\pi_{k-1}\). For this path, \(\pi_{0}\) denotes the
CoT-SFT checkpoint. We also create a frozen copy of \(\pi_{k-1}\) as the
reference model \(\pi_{\mathrm{ref},k}\). This reference model remains
fixed throughout the round.

The chosen and rejected responses may have different lengths. To reduce the
effect of response length, we replace the total sequence log-probability in
Equation~\ref{eq:sequence_logprob} with the average log-probability over all
response tokens:
\begin{equation}
    \ell_{\pi}(y \mid x)
    =
    \frac{1}{|y|}
    \sum_{t=1}^{|y|}
    \log \pi
    \left(
        y_t \mid x, y_{<t}
    \right).
    \label{eq:average_logprob}
\end{equation}

Using this length-normalized score, the DPO component of L-RPO for round \(k\)
is
\begin{equation}
\begin{aligned}
    \mathcal{L}_{\mathrm{LDPO}}^{(k)}
    =
    -\mathbb{E}_{(x,y^{+},y^{-}) \sim
    \mathcal{D}_{\mathrm{pref}}^{(k)}}
    \log \sigma
    \Bigg(
    \beta \Big[
        &\ell_{\pi_{\theta,k}}(y^{+} \mid x)
        - \ell_{\pi_{\mathrm{ref},k}}(y^{+} \mid x) \\
        &-
        \left(
        \ell_{\pi_{\theta,k}}(y^{-} \mid x)
        - \ell_{\pi_{\mathrm{ref},k}}(y^{-} \mid x)
        \right)
    \Big]
    \Bigg).
    \label{eq:length_normalized_dpo}
\end{aligned}
\end{equation}

DPO optimizes the relative preference between the chosen and rejected
responses. However, it does not guarantee that the likelihood of the chosen
response itself will increase. The chosen response can become relatively more
likely simply because the likelihood of the rejected response decreases more.
We therefore add a supervised loss that directly increases the likelihood of
the chosen response:
\begin{equation}
    \mathcal{L}_{\mathrm{chosen}}^{(k)}
    =
    -\mathbb{E}_{(x,y^{+},y^{-}) \sim
    \mathcal{D}_{\mathrm{pref}}^{(k)}}
    \ell_{\pi_{\theta,k}}(y^{+} \mid x).
    \label{eq:chosen_loss}
\end{equation}

We combine the length-normalized DPO loss and the chosen-response loss to
obtain the L-RPO objective:
\begin{equation}
    \mathcal{L}_{\mathrm{L\text{-}RPO}}^{(k)}
    =
    \mathcal{L}_{\mathrm{LDPO}}^{(k)}
    +
    \alpha
    \mathcal{L}_{\mathrm{chosen}}^{(k)}.
    \label{eq:lrpo}
\end{equation}

For example, we first train L-RPO on
\(\mathcal{D}_{\mathrm{pref}}^{(1)}\). We then use the resulting model to
generate new responses and collect its remaining errors. These errors are
paired with the gold responses to construct
\(\mathcal{D}_{\mathrm{pref}}^{(2)}\), and the first-round model is used to
initialize the second L-RPO round. As shown in
Figure~\ref{fig:architecture}, this policy-dependent data refresh forms
iterative L-RPO.
Throughout preference optimization, we update only the LoRA adapter of the model.

\section{Experimental Setup}

\subsection{Datasets and Evaluation Metrics}
The direct-answer SFT dataset includes 35,000 examples from TableInstruct and 70,000 examples created from MammoTab V2. After removing examples with overly long lengths, 104,899 examples remain.
The CoT-SFT dataset also combines MammoTab V2 and TableInstruct. After filtering the data, compressing the reasoning traces, and removing examples that exceed the length limit, 35,617 examples remain.
For validation, we use a fixed set of 3,750 processed MammoTab tables drawn from the original training set and sample one target mention from each table. These tables are held out and are not used for training. At every training stage, we select the checkpoint with the best performance on this validation set.
For final evaluation, we use the full TableInstruct entity-linking test set and MammoTab V2 evaluation set. MammoTab V2 evaluation set contains 9,741 entity mentions from 511 tables.
Our main evaluation metric is exact-match accuracy. A prediction is counted as correct only when its final answer exactly matches the gold answer.
For reasoning models, we also report the complete-rationale rate. It is the percentage of outputs with a complete
\texttt{<think>...\allowbreak</think>} block. Table~\ref{tab:dataset-statistics} summarizes the dataset statistics throughout data construction and evaluation.

\begin{table}[t]
  \centering
  \caption{Dataset statistics throughout data construction and evaluation.}
  \label{tab:dataset-statistics}
  \small
  \begin{tabular}{@{}lllr@{}}
    \toprule
    Stage & Data Source  & Count \\
    \midrule
    Original training data & TableInstruct entity linking  & 1,264,217 \\
    Sampled training data  & TableInstruct entity linking  & 168,764 \\
    Original training data & MammoTab V2 & 29,107,433 \\
    Sampled training data & MammoTab V2 & 838,182 \\
    \midrule
    Direct-answer SFT selection & TableInstruct  & 35,000 \\
    Direct-answer SFT selection & MammoTab V2 & 70,000 \\
    Direct-answer SFT, merged & TableInstruct + MammoTab V2 & 105,000 \\
    Direct-answer SFT, after length check & TableInstruct + MammoTab V2  & 104,899 \\
    \midrule
    CoT-SFT, Rationale generation & DeepSeek-V4-Pro + GPT-5.2 (Reasoning) & 36,276 \\
    CoT-SFT, After quality filtering & DeepSeek-V4-Pro + GPT-5.2 (Reasoning)  & 35,641 \\
    CoT-SFT, After final length check & DeepSeek-V4-Pro + GPT-5.2 (Reasoning)  & 35,617 \\
    \midrule
    L-RPO round 1, generated & DeepSeek-V4-Pro & 4,052 \\
    L-RPO round 2, generated & DeepSeek-V4-Pro & 2,541 \\
    \midrule
    Validation data & Held-out subset of the MammoTab V2 training set & 3,750 \\
    Test data & MammoTab V2 & 9,741 \\
    \bottomrule
  \end{tabular}
\end{table}

\subsection{Baselines and Ablations}
We study two training paths. The direct-answer path starts with SFT and then uses two rounds of iterative DPO. 
The reasoning path adds CoT-SFT and then uses two rounds of iterative L-RPO. In both paths, the second preference-learning round uses new data generated by the model from the first round.
For baseline comparison, we use the scores reported on the official MammoTab V2 page: 0.86 for TableLlama~\cite{zhang2023tablellama}, 0.31 for
TURL~\cite{deng2020turl}, and 0.62 for Avogadro 2023~\cite{avogadro2023estimating}. 
Since each target mention receives one prediction, our exact-match accuracy is the CEA score used in this comparison.

\subsection{Implementation Details}
We use Llama-3.1-8B as the backbone. We train it with LoRA using rank 8, a scaling factor of 16, and no LoRA dropout. 
The embedding and normalization parameters are also trainable. Each stage updates the LoRA adapter from the previous stage. The base model stays frozen.
Direct-answer SFT runs for two epochs with a learning rate of
$2\times10^{-5}$ and a maximum sequence length of 2,048. CoT-SFT runs for one epoch with the same learning rate and a maximum sequence length of 3,072. 
Both stages use bfloat16 and a cosine learning-rate schedule with a 3\% warm-up.

Each L-RPO round runs for one epoch. We use a learning rate of
$1.5\times10^{-6}$, $\beta=0.3$, and $\alpha=0.5$. The maximum sequence length is 2,560. 
We remove a preference pair if the ratio between its chosen and rejected response lengths is greater than 6. 
At the start of each round, the trainable adapter and the reference adapter have the same weights. 
Only the trainable adapter is updated. The direct-answer DPO rounds run for two epochs with a learning rate of $5\times10^{-6}$.
Training uses H100 80GB GPUs. We choose the best checkpoint for each model and ablation on the validation set. 
For L-RPO, the selection score is the sum of exact-match accuracy and the complete-rationale rate. Direct-answer models can generate up to 128 new tokens. 
Reasoning models can generate up to 768 new tokens. 
All ablation studies and model-training experiments use the same data-processing and preprocessing code, as well as the same evaluation pipeline. All reported experimental results are averaged over five repeated evaluation runs.

\section{Results and Discussion}
\subsection{Main Results}
Table~\ref{tab:main_results} reports results on the TableInstruct
entity-linking subset and MammoTab V2.  In the direct-answer path, iterative
DPO improves accuracy monotonically on both datasets.  On TableInstruct,
accuracy increases from 94.35\% after SFT to 94.40\% after DPO-1 and 94.50\%
after DPO-2.  On MammoTab V2, the corresponding progression is 87.59\%,
88.16\%, and 88.20\%.  The two DPO rounds therefore provide total gains of
0.15 and 0.61 percentage points over SFT, respectively, with the refreshed
second round retaining the improvement on both datasets.

The reasoning path shows a different pattern.  On TableInstruct, CoT-SFT
achieves 92.90\% accuracy and a 98.50\% complete-rationale rate.  Iterative
L-RPO raises accuracy to 92.95\% and rationale completion to 99.55\%; the
first round reaches the highest rationale-completion rate, 99.70\%.  On
MammoTab V2, L-RPO-2 improves CoT-SFT accuracy from 79.09\% to 81.85\% and
rationale completion from 87.09\% to 91.86\%, gains of 2.76 and 4.77
percentage points.  Although both preference-learning paths improve their
respective initializations, direct-answer generation remains more accurate
than explicit reasoning on both datasets.

\begin{table}[t]
  \centering
  \caption{Results on the TableInstruct entity-linking subset and the MammoTab
  V2 evaluation set. Reasoning completeness is not applicable to direct-answer
  models.}
  \label{tab:main_results}
  \small
  \setlength{\tabcolsep}{4.5pt}
  \renewcommand{\arraystretch}{1.08}
  \begin{tabular}{llcccc}
    \toprule
    \multirow{2}{*}{\textbf{Training Type}} &
    \multirow{2}{*}{\textbf{Model Stage}} &
    \multicolumn{2}{c}{\textbf{TableInstruct EL}} &
    \multicolumn{2}{c}{\textbf{MammoTab V2}} \\
    \cmidrule(lr){3-4}
    \cmidrule(lr){5-6}
    & & \textbf{Accuracy} & \makecell{\textbf{Reasoning}\\\textbf{Complete}}
      & \textbf{Accuracy} & \makecell{\textbf{Reasoning}\\\textbf{Complete}} \\
    \midrule

    Pretrained
      & Base Model (Llama-3.1-8B)
      & 58.60\% & n/a
      & 45.78\% & n/a \\
    \midrule

    \multirow{3}{*}{\makecell[l]{Direct-Answer Training}}
      & Direct-Answer SFT
      & 94.35\% & n/a
      & 87.59\% & n/a \\
      & Iterative DPO Round 1
      & 94.40\% & n/a
      & 88.16\% & n/a \\
      & Iterative DPO Round 2
      & \textbf{94.50\%} & n/a
      & \textbf{88.20\%} & n/a \\
    \midrule

    \multirow{3}{*}{\makecell[l]{Reasoning-Based Training}}
      & CoT-SFT
      & 92.90\% & 98.50\%
      & 79.09\% & 87.09\% \\
      & L-RPO Round 1
      & 92.90\% & \textbf{99.70\%}
      & 79.37\% & 89.63\% \\
      & L-RPO Round 2
      & \textbf{92.95\%} & 99.55\%
      & \textbf{81.85\%} & \textbf{91.86\%} \\
    \bottomrule
  \end{tabular}
\end{table}

Table~\ref{tab:mammotabv2_comparison} compares the best model with the public
results listed for the same 511-table, 9,741-mention MammoTab V2 evaluation
set~\cite{mammotabV2evaluation}.  Iterative DPO round 2 reaches 0.882 CEA,
surpassing the previous best listed score of 0.86 by 2.2 percentage points.

\begin{table}[t]
  \centering
  \caption{CEA comparison on the MammoTab V2 evaluation set.  Public baseline
  scores are reported by MammoTab V2.}
  \label{tab:mammotabv2_comparison}
  \begin{tabular}{lr}
    \hline
    Approach & CEA \\
    \hline
    TURL (Deng et al.)~\cite{deng2020turl} & 0.31 \\
    Avogadro 2023~\cite{avogadro2023estimating} & 0.62 \\
    TableLlama (Zhang et al.)~\cite{zhang2023tablellama} & 0.86 \\
    Iterative DPO round 2 (ours) & \textbf{0.882} \\
    \hline
  \end{tabular}
\end{table}

\subsection{Ablation Study}
Table~\ref{tab:loss_ablation} evaluates the effects of completion-length normalization and the chosen-response loss in both L-RPO rounds. Removing length normalization consistently reduces answer accuracy and rationale completeness. In round 1, accuracy decreases from 79.37\% to 69.53\%, while rationale completeness drops from 89.63\% to 77.06\%. Relative to the CoT-SFT initialization on identical prompts, this variant corrects 583 previously incorrect predictions but regresses 1,514 previously correct ones. The same pattern persists in round 2, where removing length normalization reduces accuracy from 81.85\% to 79.42\% and rationale completeness from 91.86\% to 87.54\%. Restoring length normalization at this stage corrects 714 predictions while regressing 477. These consistent degradations support normalizing the policy and reference log-probabilities when the chosen and rejected responses differ in length.

The chosen-response loss is similarly important. Removing it in round 1 reduces accuracy from 79.37\% to 68.52\% and rationale completeness from 89.63\% to 74.30\%. Compared with the CoT-SFT initialization on identical prompts, this update corrects 532 errors but regresses 1,561 correct predictions. In round 2, removing the chosen-response loss lowers accuracy from 81.85\% to 78.13\% and rationale completeness from 91.86\% to 84.11\%. The degradation across both rounds is consistent with the chosen-response loss anchoring the model to high-likelihood responses with valid reasoning and well-formed output structure.

The effects of both components are substantially larger in round 1 than in round 2, suggesting that they are particularly important for stabilizing the initial preference-optimization update.

\begin{table}[t]
  \centering
  \caption{Ablation of the L-RPO loss components on the complete evaluation
  set. LN denotes completion-length normalization, and the chosen loss denotes the likelihood regularization applied to the chosen response.}
  \label{tab:loss_ablation}
  \small
  \setlength{\tabcolsep}{5pt}
  \renewcommand{\arraystretch}{1.08}
  \begin{tabular}{clcccc}
    \toprule
    \textbf{Round} &
    \textbf{Variant} &
    \textbf{LN} &
    \makecell{\textbf{Chosen Loss}} &
    \textbf{Accuracy} &
    \makecell{\textbf{Reasoning}\\\textbf{Complete}} \\
    \midrule

    \multirow{3}{*}{2}
      & Full
      & Yes & Yes
      & \textbf{81.85\%}
      & \textbf{91.86\%} \\
      & Without LN
      & No & Yes
      & 79.42\%
      & 87.54\% \\
      & Without chosen loss
      & Yes & No
      & 78.13\%
      & 84.11\% \\
    \midrule

    \multirow{3}{*}{1}
      & Full
      & Yes & Yes
      & 79.37\%
      & 89.63\% \\
      & Without LN
      & No & Yes
      & 69.53\%
      & 77.06\% \\
      & Without chosen loss
      & Yes & No
      & 68.52\%
      & 74.30\% \\
    \bottomrule
  \end{tabular}
\end{table}

\subsection{Discussion and Limitations}

Table~\ref{tab:transitions} shows that the aggregate gains arise from a combination of corrected and regressed predictions. Relative to CoT-SFT, the full two-round L-RPO model corrects 817 previously incorrect predictions while causing 548 previously correct predictions to fail, yielding a net gain of 269 correct predictions. Compared with L-RPO-2 without length normalization, the full model similarly achieves a net gain of 237 predictions. Rationale completeness exhibits the same trend. From CoT-SFT to full L-RPO-2, 869 outputs gain a complete reasoning block, whereas 404 lose one. Adding length normalization to L-RPO-2 produces 955 gains and 534 losses relative to the no-LN variant. By contrast, the first-round update without length normalization yields only 693 gains but 1,670 losses in rationale completeness relative to CoT-SFT, indicating that length normalization is important for stabilizing the reasoning format.

\begin{table}[t]
  \centering
  \caption{Paired correctness analysis on the same $9{,}741$ test prompts.
  Corrected and regressed denote wrong-to-correct and correct-to-wrong
  transitions, respectively. Net gain is computed as corrected minus
  regressed.}
  \label{tab:transitions}
  \small
  \setlength{\tabcolsep}{4.2pt}
  \renewcommand{\arraystretch}{1.08}
  \begin{tabular}{llrrrrr}
    \toprule
    \multirow{2}{*}{\textbf{Comparison Type}} &
    \multirow{2}{*}{\textbf{Transition}} &
    \multicolumn{4}{c}{\textbf{Paired Outcome}} &
    \multirow{2}{*}{\makecell{\textbf{Net}\\\textbf{Gain}}} \\
    \cmidrule(lr){3-6}
    & &
    \makecell{\textbf{Both}\\\textbf{Correct}} &
    \textbf{Corrected} &
    \textbf{Regressed} &
    \makecell{\textbf{Both}\\\textbf{Wrong}} & \\
    \midrule

    Full training
    & CoT-SFT $\rightarrow$ L-RPO-2
    & 7,156 & 817 & 548 & 1,220 & $+269$ \\

    Component effect
    & L-RPO-2 w/o LN $\rightarrow$ full
    & 7,259 & 714 & 477 & 1,291 & $+237$ \\

    \midrule

    \multirow{2}{*}{Ablated training}
    & CoT-SFT $\rightarrow$ L-RPO-1 w/o LN
    & 6,190 & 583 & 1,514 & 1,454 & $-931$ \\

    & CoT-SFT $\rightarrow$ L-RPO-1 w/o RPO
    & 6,143 & 532 & 1,561 & 1,505 & $-1{,}029$ \\

    \bottomrule
  \end{tabular}
\end{table}

Several limitations remain.  First, every training instance in the final
dataset contains the gold entity among its retrieved candidates; consequently,
the dataset contains no empty or NIL targets.  The current model is therefore
not trained to handle cases in which candidate retrieval fails to return the
correct entity.  Extending the model to generate an explicit NIL prediction
when none of the retrieved candidates is appropriate would broaden its coverage
of real-world table entity-linking scenarios and could further improve its
robustness and overall performance.

Second, the training rationales are generated by teacher models,
which makes the rationale-generation process difficult to reproduce exactly.
To mitigate this limitation, we release all training data and code in our
repository, including the complete set of rationales used in this work.  These
resources enable other researchers to reproduce our training setup and results
without regenerating the rationale supervision, and provide a foundation for
further research.

Third, our iterative DPO and L-RPO pipelines are evaluated for only two rounds. Although both rounds consistently improve upon the single-round baseline, it remains unclear whether additional iterations would yield further gains or eventually lead to performance saturation. Exploring longer iteration horizons and identifying the potential saturation point are left for future work.

\section{Conclusion}
We introduced TELLER, a dual-path framework for candidate-conditioned table
entity linking that learns iteratively from model errors and reasoning.  Its offline pipeline
retrieves and ranks Wikidata candidates, while a compact prompt preserves the
target cell, relevant row and column context, table headers, and the caption
and removes the remaining table cells.  In the direct-answer path, each DPO
round refreshes its preference data with the residual errors of the updated
model.  Across two rounds, accuracy improves from 94.35\% to 94.50\% on
TableInstruct and from 87.59\% to 88.20\% on MammoTab V2.  The latter result
corresponds to a CEA score of 0.882, surpassing the highest score currently
listed on the public MammoTab V2 evaluation page.

In the reasoning path, filtered and compressed rationales provide supervision
for CoT-SFT, followed by iterative L-RPO.  L-RPO combines length-normalized
preference scores with a chosen-response likelihood loss, and its second round
replaces earlier rejected outputs with the residual reasoning errors of
first round.  The final model improves CoT-SFT accuracy from 92.90\% to 92.95\% on
TableInstruct and from 79.09\% to 81.85\% on MammoTab V2, while increasing the
complete-rationale rate on MammoTab V2 from 87.09\% to 91.86\%.  Paired error
analysis shows significantly more corrected than regressed predictions, and
the ablations provide further evidence that both length normalization and
chosen-response regularization are important for maintaining answer accuracy
and complete reasoning.  Nevertheless, the direct-answer models remain more
accurate on both datasets, indicating a persistent trade-off between explicit
reasoning and entity-selection accuracy. Future work should expand and diversify the rationale data,
integrate table data from additional domains to improve cross-domain
generalization, and extend the model with a NIL option for cases in which none
of the retrieved candidates is correct. 

\paragraph*{Supplemental Material Statement:}
Source code and dataset are available at \href{https://github.com/787602925/TELLER} {this github repository}~\cite{TELLER2026}.

\section*{Declaration on Generative AI}
DeepSeek-V4-Pro and GPT-5.2 (Reasoning) were used as teacher models to generate training rationales, as described in the methodology. No generative AI tools were used for writing, editing, or creating figures, texts and other content in this work.

\bibliography{reference}

\appendix

\section{Data Examples}
\label{app:data-examples}

This appendix presents representative serialized records from the datasets
used in the two training paths.  Line wrapping is added only for readability.

\subsection{TableInstruct-Style Entity-Linking Instance}
\label{appendix:el-example}

The following record illustrates the common instruction format used for a
TableInstruct example and a processed MammoTab example.

\begin{lstlisting}[
  basicstyle=\ttfamily\scriptsize,
  columns=fullflexible,
  keepspaces=true,
  literate={•}{{\textbullet}}1
]
{
 'instruction': 'This is an entity linking task. The goal for this task is to link the selected entity mention in the table cells to the entity in the knowledge base. You will be given a list of referent entities, with each one composed of an entity name, its description and its type. Please choose the correct one from the referent entity candidates. Note that the Wikipedia page, Wikipedia section and table caption (if any) provide important information for choosing the correct referent entity.',
 'input_seg': '[TLE] The Wikipedia page is about Gay Left. The Wikipedia section is about The Collective. [TAB] col: | issue/name | issue 1 autumn 1975 | issue 2 spring 1976 | issue 4 summer 1977 | issue 5 winter 1977/8 | issue 6 summer 1978 | issue 7 winter 1978/9 | issue 8 summer 1979 | issue 9 winter 1979/80 | row 1: Keith Birch [SEP] row 2: Gregg Blachford [SEP] row 3: Bob Cant [SEP] row 4: | Emmanuel Cooper | • | • | • | • | • | • | • | • | [SEP] row 5: Ross Irwin [SEP] row 6: Randall Kincaid [SEP] row 7: Angus Suttie [SEP] row 8: Jeffrey Weeks [SEP] row 9: Nigel Young [SEP] row 10: Derek Cohen',
 'question': 'The selected entity mention in the table cell is: Emmanuel Cooper. The column name for \'Emmanuel Cooper\' is issue/name. The referent entity candidates are: <Richard Dyer [DESCRIPTION] researcher ORCID ID = 0000-0002-0090-7580 [TYPE] None>, <Richard Dyer-Bennet [DESCRIPTION] American musician [TYPE] owl#Thing>, <Richard Dyer Mudd. [DESCRIPTION] scientific article [TYPE] None>, <Emmanuel Cooper OBE, 1938-2012 [DESCRIPTION] edition; published in 2013 [TYPE] None>, <Richard Dyer [DESCRIPTION] fencer [TYPE] Athlete>, <Emmanuel Cooper [DESCRIPTION] book (work) [TYPE] None>, <Jeffrey Weeks [DESCRIPTION] American mathematician [TYPE] owl#Thing>, <Angus Suttie [DESCRIPTION] book (work) [TYPE] None>, <Angus Suttie [DESCRIPTION] edition; published in 2018 [TYPE] None>, <Jeffrey Weeks [DESCRIPTION] British historian and sociologist [TYPE] owl#Thing>, <Richard Dyer [DESCRIPTION] born 1634 [TYPE] None>, <Emmanuel Cooper [DESCRIPTION] Studio potter, writer [TYPE] owl#Thing>, <Richard Dyer [DESCRIPTION] British academic [TYPE] owl#Thing>, <Emmanuel Cooper OBE, 1938-2012 [DESCRIPTION] book (work) [TYPE] None>, <Angus Suttie [DESCRIPTION] British artist (1946-1993) [TYPE] owl#Thing>, <Jeffrey Weeks and the History of Sexuality [DESCRIPTION] None [TYPE] None>, <Emmanuel Cooper [DESCRIPTION] edition; published in 1992 [TYPE] None>, <Richard Dyer [DESCRIPTION] None [TYPE] None>, <Richard Dyer [DESCRIPTION] Montserratian footballer [TYPE] None>, <Jeffrey Weeks [DESCRIPTION] Wikimedia disambiguation page [TYPE] None> What is the correct referent entity for the entity mention \'Emmanuel Cooper\'?',
 'output': '<Emmanuel Cooper [DESCRIPTION] Studio potter, writer [TYPE] owl#Thing>'
}
\end{lstlisting}

\subsection{Preference Instance after CoT-SFT}
\label{appendix:cot-pref-example}

This example contains the prompt, the chosen and rejected reasoning-based
responses, the gold entity, and the model prediction used to construct the
pair.  The candidate list is abbreviated in the displayed record.

\begin{lstlisting}[
  basicstyle=\ttfamily\scriptsize,
  columns=fullflexible,
  keepspaces=true,
  literate={⋅}{{$\cdot$}}1
]
{
 'idx': 35,
 'prompt': '### Instruction:\nentity linking task. choose only the correct one from the referent entity candidates. In the Input below, the table content only contains the caption (if any), all column headers, and the cells from the same row and same column as the selected entity mention.\n\n### Input:\n[TLE] The Wikipedia page is about Energy density. The Wikipedia section is about List of material energy densities. The table caption is about Energy released by electrochemical reactions or other means. [TAB] col: | Material | Specific energy (MJ/kg) | Energy density (MJ/L) | Specific energy (W⋅h/kg) | Energy density (W⋅h/L) | Comment | row 1: Battery, zinc-air [SEP] row 2: Silicon (phase change) [SEP] row 3: | Strontium bromide hydrate | 0.814 A. Fopah-Lele, J. G. Tamba A review on the use of as a potential material for low temperature energy storage systems and building applications, Solar Energy Materials and Solar Cells 164 175-84 (2017). | 1.93 |  | 628 | Thermal energy of phase change at | [SEP] row 4: Liquid nitrogen [SEP] row 5: Compressed air at 30 MPa [SEP] row 6: Latent heat of fusion of ice (thermal) [SEP] row 7: Lithium metal battery [SEP] row 8: Lithium-ion battery [SEP] row 9: Lithium-ion battery with silicon nanowire anodes [SEP] row 10: Flywheel\n\n### Question:\nThe selected entity mention in the table cell is: Strontium bromide hydrate. The column name for \'Strontium bromide hydrate\' is Material. The referent entity candidates are: ...(20 candidates) What is the correct referent entity for the entity mention \'Strontium bromide hydrate\'?\n\nReasoning requirements:\n1) Use context evidence only (page/section/caption + row/column + candidate [DESCRIPTION]/[TYPE]).\n2) Keep reasoning concise; do NOT restate the full table/question/candidate list.\n3) Do NOT copy/paste candidate lists from the input; only cite minimal distinguishing evidence.\n4) The reasoning should support the same final entity you output after </think>.\n5) Reasoning is mandatory: include non-empty reasoning content before the final entity; do not output answer-only.\n\nAfter reasoning, output exactly one referent entity in candidate format (e.g. <EntityName [DESCRIPTION] ... [TYPE] ...>).\n\n### Response:',
 'accept': '<think>...</think>\n<strontium bromide [DESCRIPTION] chemical compound [TYPE] type of chemical entity>',
 'reject': '<think>...</think>\n<strontium bromide hexahydrate [DESCRIPTION] chemical compound [TYPE] type of chemical entity>',
 'gold': '<strontium bromide [DESCRIPTION] chemical compound [TYPE] type of chemical entity>',
 'model_prediction': '<strontium bromide hexahydrate [DESC] chemical compound [TYPE] type of chemical entity>'
}
\end{lstlisting}

\subsection{Preference Instance after Direct-Answer SFT}
\label{appendix:direct-answer-pref-example}

The following record illustrates a direct-answer preference pair constructed
from an incorrect SFT-model prediction.

\begin{lstlisting}[
  basicstyle=\ttfamily\scriptsize,
  columns=fullflexible,
  keepspaces=true,
  literate={–}{{--}}1
]
{
 'idx': 82,
 'prompt': '### Instruction:\nentity linking task. choose only the correct one from the referent entity candidates. In the Input below, the table content only contains the caption (if any), all column headers, and the cells from the same row and same column as the selected entity mention.\n\n### Input:\n[TLE] The Wikipedia page is about List of Clyde F.C. seasons. The Wikipedia section is about Seasons. [TAB] col: | Season | League | League | Scottish Cup | League Cup | Other | Top league scorer | Top league scorer | row 8: Division A [SEP] row 9: Division A [SEP] row 10: Division A [SEP] row 11: Division One [SEP] row 12: | 1956-57 | Division Two | 1st | Quarter-final | Semi-final | Glasgow Cup runners-up | Basil Keogh | 36 | [SEP] row 13: Division One [SEP] row 14: Division One [SEP] row 15: Division One [SEP] row 16: Division One [SEP] row 17: Division Two\n\n### Question:\nThe selected entity mention in the table cell is: Division Two. The column name for \'Division Two\' is League. The referent entity candidates are: <Division Two [DESCRIPTION] Wikimedia disambiguation page [TYPE] Wikimedia disambiguation page>, <BAFA NL Division Two 2016 [DESCRIPTION] None [TYPE] sports season>, <South Division Two [DESCRIPTION] Wikimedia disambiguation page [TYPE] Wikimedia disambiguation page>, <BAFA NL Division Two 2015 [DESCRIPTION] None [TYPE] sports season>, <NCL Division Two [DESCRIPTION] None [TYPE] None>, <1911–12 Scottish Division Two [DESCRIPTION] football league season [TYPE] sports season>, <Sapphire Series Division Two 2017 [DESCRIPTION] None [TYPE] sports season>, <1956–57 Scottish Division Two [DESCRIPTION] football league season [TYPE] sports season>, <1972–73 Scottish Second Division [DESCRIPTION] football league season [TYPE] sports season>, <2000–01 National Division Two [DESCRIPTION] sports season [TYPE] sports competition>, <North Division Two [DESCRIPTION] football league [TYPE] None>, <2006–07 WRU Division Two West [DESCRIPTION] None [TYPE] sports season>, <1966–67 Scottish Division Two [DESCRIPTION] football league season [TYPE] sports season>, <Barbados Division Two [DESCRIPTION] football league [TYPE] sports league>, <2024 National Division Two [DESCRIPTION] None [TYPE] sports season>, <South Division Two [DESCRIPTION] football league [TYPE] None>, <Leinster League Division Two [DESCRIPTION] None [TYPE] None>, <Division Two League, Ghana [DESCRIPTION] None [TYPE] association football league>, <1928–29 Scottish Division Two [DESCRIPTION] football league season [TYPE] sports season>, <2006–07 WRU Division Two East [DESCRIPTION] None [TYPE] sports season>, <WRU Division Two North [DESCRIPTION] None [TYPE] sports division> What is the correct referent entity for the entity mention \'Division Two\'?\n\n### Response:',
 'accept': '<1972–73 Scottish Second Division [DESCRIPTION] football league season [TYPE] sports season>',
 'reject': '<1956–57 Scottish Division Two [DESCRIPTION] football league season [TYPE] sports season>',
 'gold': '<1972–73 Scottish Second Division [DESCRIPTION] football league season [TYPE] sports season>',
 'model_prediction': '<1956–57 Scottish Division Two [DESC] football league season [TYPE] sports season>'
}
\end{lstlisting}

\end{document}